\documentclass[runningheads]{llncs}
\usepackage{graphicx}
\usepackage{amsmath}
\usepackage{url}
\usepackage{tikz}

\usepackage[chapter]{algorithm}
\usepackage{algpseudocode}
\usepackage{forarray}
\usepackage{xspace}
\usepackage{amssymb}

\newcommand{\asp}[1]{\mbox{$\mathtt{#1}$}}
\DeclareMathOperator{\codeif}{\mathtt{:-} }

\newcommand{\answerset}[1]{
  \lbrace$\ForEach{;}{\ifnum\thislevelcount=1 \else ,$ $ \fi \asp{\thislevelitem}}{#1}$\rbrace
}

\usepackage{tikz}
\usetikzlibrary{shapes,chains,arrows.meta,calc,decorations.markings,math,arrows.meta}

\begin{document}

\title{The ILASP System for Inductive Learning of Answer Set Programs}

\author{Mark Law\inst{1,2} \and Alessandra Russo\inst{2} \and Krysia Broda\inst{2}}

\authorrunning{Mark Law et al.}

\institute{ILASP Limited, UK\\ \email{mark@ilasp.com}\\ \and Department of
Computing, Imperial College London, UK\\ \email{$\lbrace$mark.law09, a.russo,
k.broda$\rbrace$@imperial.ac.uk}}

\maketitle

\begin{abstract}
  The goal of Inductive Logic Programming (ILP) is to learn a program that
explains a set of examples in the context of some pre-existing background
knowledge. Until recently, most research on ILP targeted learning Prolog
programs. Our own ILASP system instead learns Answer Set Programs, including
normal rules, choice rules and hard and weak constraints. Learning such
expressive programs widens the applicability of ILP considerably; for example,
enabling preference learning, learning common-sense knowledge, including
defaults and exceptions, and learning non-deterministic theories. In this
paper, we first give a general overview of ILASP's learning framework and its
capabilities. This is followed by a comprehensive summary of the evolution of
the ILASP system, presenting the strengths and weaknesses of each version, with
a particular emphasis on scalability.

\end{abstract}

%

\section{Introduction}

The ability to declaratively specify real-world problems and efficiently
generate solutions from such specifications is of particular interest in both
academia and industry~\cite{falkner2018industrial,erdem2016applications}.
A typical paradigm is Answer Set Programming
(ASP)~\cite{gelfond1988,brewka2011answer}, which allows a problem to be
described in terms of its specification, rather than requiring a user to
define an algorithm to solve the problem.
Its solvers are capable of constructing solutions from the specifications alone
and, where needed, ranking solutions according to optimisation criteria.
The interpretable nature of the ASP language also enables the generation of
explanations, which is particularly relevant in AI-driven applications.
Due to its rich language and efficient solving, ASP has been applied to a wide
range of classical areas in AI -- including planning, scheduling and diagnosis
-- and is increasingly being applied in industry~\cite{erdem2016applications};
for example, in decision support systems, automated product configuration and
configuration of safety systems~\cite{falkner2018industrial}.
On the other hand, developing ASP specifications of real-world problems can be
a difficult task for non-experts.
Furthermore, the dynamic nature of the contexts in which real-world AI
applications tend to operate can require the ASP specification of a problem to
be regularly updated or revised.
To widen the dissemination of ASP in practice, it is therefore crucial to
develop methods that can automatically learn ASP specifications from examples
of (partial) solutions to real-world problems.
Such learning mechanisms could also provide ways for automatically evolving and
revising ASP specifications in dynamic environments.

Within the last few years, we have addressed this problem and developed a novel
system, called \emph{Inductive Learning of Answer Set Programs}
(ILASP)~\cite{ILASP_system}.
The theoretical framework underpinning the ILASP system differs from
conventional approaches for Inductive Logic Programming (ILP), which are mainly
focused on learning Prolog programs.
Due to the declarative nature of ASP, the learning process in ILASP primarily
targets learning the logical specification of a problem, rather than the
procedure for solving that problem.
Secondly, programs learned by ILASP can include extra types of rules that are
not available in Prolog, such as choice rules and hard and weak constraints.
Enabling the learning of these extra rules has opened up new applications,
which were previously out of scope for ILP systems; for instance, learning weak
constraints allows ILASP to learn a user's preferences from examples of which
solutions the user prefers~\cite{ICLP15}.

ILASP's learning framework has been proved to generalise existing frameworks
and systems for learning ASP programs~\cite{AIJ17}, such as the brave learning
framework~\cite{Sakama2009}, adopted by almost all previous systems (e.g.\
XHAIL~\cite{ray2009nonmonotonic}, ASPAL~\cite{Corapi2012}, ILED~\cite{ILED},
RASPAL~\cite{raspal}), and the less common cautious learning
framework~\cite{Sakama2009}.
Brave systems require the examples to be covered in at least one answer set of
the learned program, whereas cautious systems find a program which covers the
examples in every answer set.
We showed in~\cite{AIJ17} that some ASP programs cannot be learned with either
a brave or a cautious approach, and that to learn ASP programs in general, a
combination of both brave and cautious reasoning is required.
ILASP's learning framework enables this combination, and is capable of learning
the full class of ASP programs~\cite{AIJ17}.
ILASP's generality has allowed it to be applied to a wide range of
applications, including event detection~\cite{ACS18}, preference
learning~\cite{ICLP15}, natural language
understanding~\cite{chabierski2017machine}, learning game
rules~\cite{cropper2019inductive}, grammar induction~\cite{law2019representing}
and automata induction~\cite{danielf}.

In this paper we give an introduction to ILASP's learning framework and its
capabilities, demonstrating the various types of examples that ILASP can learn
from, and describe the evolution of the ILASP system.
This evolution has been driven by a need for efficiency with respect to various
dimensions, including handling noisy data, large numbers of examples and large
search spaces. We discuss the strengths and weaknesses of each variation of the
ILASP system, explaining how each system improves on the efficiency of its
predecessor. We conclude with a discussion of recent developments and future
research directions.

\section{Components of a Learning Task}

ILASP is used to solve a \emph{learning task}, which consists of three main
components: the background knowledge, the mode bias and the examples.
The \emph{background knowledge} $B$ is an ASP program, which describes a set of
concepts that are already known before learning. ILASP accepts a
subset\footnote{For a formal definition of this subset, please see the ILASP
manual (\url{http://www.ilasp.com/manual}).} of ASP, consisting of normal rules,
choice rules and hard and weak constraints. We use the term \emph{rule} to
refer to any of these four components.

The \emph{mode bias} $M$ (often called a language bias) is used to express the
ASP programs that can be learned; for example, it specifies which predicates
may be used in the head/body of learned rules, and how they may be used
together. From $M$, it is possible to construct a (finite) set of rules $S_M$
called the \emph{rule space}\footnote{In other literature, the rule space is
often called the \emph{hypothesis space}.}, that contains every rule that is
compatible with $M$. The power set of $S_M$, $\mathbb{P}(S_M)$, is called the
\emph{program space}, and contains the set of all ASP programs that can be
learned.

The \emph{examples} $E$ describe a set of semantic properties that the learned
program should satisfy. When the semantic property of an individual example
$e\in E$ is satisfied, we say that $e$ is \emph{covered}. The goal of an ILP
system, such as ILASP, is to find a program (often called a hypothesis) $H \in
\mathbb{P}(S_M)$ such that $B\cup H$ (the combination of this program with the
background knowledge) covers every example in $E$. Many ILP systems (including
ILASP) follow the principle of Occam's Razor, that the simplest solution should
be preferred, and therefore search for an \emph{optimal} program, which is the
shortest in terms of the number of literals.

Many ILP systems learn from (positive and negative) examples of atoms which
should be true or false. This is because many ILP systems are targeted at
learning Prolog programs, where the main ``output'' of a program is a query of
a single atom. In ASP, the main ``output'' of a program is a set of answer
sets. For this reason, ILASP learns from positive and negative examples of
(partial) interpretations, which should or should not (respectively) be an
answer set of the learned program. These examples are sufficient to learn any
ASP program consisting of normal rules, choice rules and hard constraints (up
to strong equivalence)~\cite{AIJ17}; however, it is not possible to learn weak
constraints using only positive and negative examples, because they can only
specify what should (or should not) be an answer set. Weak constraints do not
have any effect on what is or is not an answer set -- they only create a
preference ordering over the answer sets. For this reason, ILASP allows a
second type of example called an \emph{ordering example}, the semantic property
of which is a preference ordering over a pair of answer sets of $B\cup H$.
Learning weak constraints corresponds to a form of \emph{preference learning}.

\subsection{Positive and Negative Examples}

Consider a very simple setting, where we want to learn a program that describes
the behaviour of three (specific) coins, by flipping them and observing which
sides they land on. We can use a very simple mode bias to describe the rules we
are allowed to learn. The predicates $\asp{heads/1}$ and $\asp{tails/1}$ are
both allowed to appear in the head with a single argument, which is either a
variable or constant of type $\asp{coin}$ (where there are three constants of
type coin in the domain: $\asp{c1}$, $\asp{c2}$ and $\asp{c3}$). In the body,
we can use three predicates (both positively and negatively): $\asp{heads/1}$,
$\asp{tails/1}$ and $\asp{coin/1}$. The mode bias expressing this language is
shown below.

\begin{multicols}{2}
\begin{verbatim}
#modeh(heads(var(coin))).
#modeh(tails(var(coin))).

#modeb(heads(var(coin))).
#modeb(tails(var(coin))).
#modeb(coin(var(coin))).

#modeh(heads(const(coin))).
#modeh(tails(const(coin))).

#constant(coin, c1).
#constant(coin, c2).
#constant(coin, c3).
\end{verbatim}
\end{multicols}

We flip the coins twice, and see the following combinations of observations:
$\answerset{heads(c1);tails(c2);heads(c3)}$,
$\answerset{heads(c1);heads(c2);tails(c3)}$.
We can encode these ``observations'' in ILASP using positive examples. Positive
examples specify properties which should hold in at least one answer set of the
learned program ($B\cup H$). The two observations are represented by the
following two examples:

\begin{verbatim}
#pos({heads(c1), tails(c2), heads(c3)},
     {tails(c1), heads(c2), tails(c3)}).

#pos({heads(c1), heads(c2), tails(c3)},
     {tails(c1), tails(c2), heads(c3)}).
\end{verbatim}

Each positive example contains two sets of ground atoms, called the inclusions
and the exclusions (respectively). For a positive example to be covered, there
must be at least one answer set of $B\cup H$ that contains all of the
inclusions and none of the exclusions. In this case, these examples mean that
there must be (at least) two answer sets, one which contains $\asp{heads(c1)}$,
$\asp{tails(c2)}$ and $\asp{heads(c3)}$, and does not contain
$\asp{tails(c1)}$, $\asp{heads(c2)}$ and $\asp{tails(c3)}$, and another answer
set which contains $\asp{heads(c1)}$, $\asp{heads(c2)}$ and $\asp{tails(c3)}$,
and does not contain $\asp{tails(c1)}$, $\asp{tails(c2)}$ and
$\asp{heads(c3)}$. Although these particular examples completely describe the
values of all coins, this does not need to be the case in general. Partial
examples allow us to represent uncertainty; for example, we could have a fourth
coin $\asp{c4}$ for which we do not know the value.

Together with the above examples, we can also give the following, very simple,
background knowledge, which defines the set of coins we have:

\begin{verbatim}
coin(c1).    coin(c2).    coin(c3).
\end{verbatim}

\noindent
If we run this task in ILASP, then ILASP returns the following solution:

\begin{verbatim}
heads(V1) :- coin(V1), not tails(V1).
tails(V1) :- coin(V1), not heads(V1).
\end{verbatim}

This program states that every coin must land on either $\asp{heads}$ or
$\asp{tails}$, but not both. Although the first coin $\asp{c1}$ has never
landed on $\asp{tails}$ in the scenarios we have observed, ILASP has
\emph{generalised} to learn first-order rules that apply to all coins, rather
than specific ground rules that only explain the specific instances we have
seen. The ability to generalise in this way is a huge advantage of ILP systems
over other forms of machine learning, because it usually means that ILP
techniques require very few examples to learn general concepts. Note that both
positive examples are required for ILASP to learn this general program. Neither
positive example on its own is sufficient because in both cases there is a
shorter program that explains the example -- the set of facts
$\lbrace\asp{heads(c1).\;tails(c2).\;heads(c3).}\rbrace$ covers the first
example and similarly the set of facts
$\lbrace\asp{heads(c1).\;heads(c2).\;tails(c3).}\rbrace$ covers the second.

It may be that after many more observations, we still have not witnessed
$\asp{c1}$ landing on $\asp{tails}$, and we could be convinced that it never
will. In this case, we can use ILASP's negative examples to specify that there
should be no answer set that contains $\asp{tails(c1)}$. This example is
expressed in ILASP as follows:

\begin{verbatim}
#neg({tails(c1)}, {}).
\end{verbatim}

\noindent
Given this extra example, ILASP learns the slightly larger program:

\begin{verbatim}
heads(V1) :- coin(V1), not tails(V1).
tails(V1) :- coin(V1), not heads(V1).
heads(c1).
\end{verbatim}

\noindent
This program states that all coins must land on either $\asp{heads}$ or
$\asp{tails}$, but not both, except for $\asp{c1}$, which can only land on
$\asp{heads}$. Note that negative examples often cause ILASP to learn programs
with rules that eliminate answer sets. In this case, the fact $\asp{heads(c1)}$
eliminates all answer sets that contain $\asp{tails(c1)}$. Negative examples
are often used to learn constraints. The constraint ``$\asp{\codeif
tails(c1).}$'' would have has the same effect; however, it is not permitted
because the mode bias does not allow constants to be used in the body of a
rule.

\subsubsection{Context-dependent Examples.}

Positive and negative examples of partial answer sets are targeted at learning
a fixed program, $B\cup H$. When ASP is used in practice, however, a program
representing a general problem definition is often combined with another set of
rules (usually just facts) describing a particular instance of the problem to
be solved.
For instance, a general program defining what it means for a graph to be
Hamiltonian (i.e.\ the general problem definition) can be combined with a set
of facts describing a particular graph (i.e.\ a problem instance).
The combined program is satisfiable if and only if the graph represented by the
set of facts is Hamiltonian. The \emph{context-dependent} behaviour of the
general Hamilton program cannot be captured by positive and negative examples
of partial answer sets. Instead, we need an extension, called a
\emph{context-dependent example}. This allows each example to come with its own
extra bit of background knowledge, called a context $C$, which applies only to
that example. It is now $B\cup H \cup C$ that has to satisfy the semantic
properties of the example, rather than $B\cup H$. In ILASP, the context of an
example is expressed by adding an extra set to the example, containing the
context.

\begin{figure}[t]
  \begin{multicols}{2}
  \begin{center}

\begin{verbatim}
#pos({}, {}, {
  node(1..4).
  edge(1, 2).
  edge(2, 3).
  edge(3, 4).
  edge(4, 1).
}).

#neg({}, {}, {
  node(1..4).
  edge(1, 2).
  edge(2, 1).
  edge(2, 3).
  edge(3, 4).
  edge(4, 2).
}).

\end{verbatim}

    \begin{tikzpicture}[scale=1.4,regular/.style={draw,ellipse, inner sep=0,minimum size=4mm}]
      \draw (1,2) node [regular] {1};
      \draw (2,2) node [regular] {2};
      \draw (1,1) node [regular] {3};
      \draw (2,1) node [regular] {4};
      \draw [->] (1.15,2) to (1.85,2);
      \draw [->] (1.90,1.90) to (1.10,1.10);
      \draw [->] (1.15,1) to (1.85,1);
      \draw [->] (1.90,1.10) to (1.10,1.90);
    \end{tikzpicture}

    \begin{verbatim}
    \end{verbatim}

    \begin{tikzpicture}[scale=1.4,regular/.style={draw,ellipse, inner sep=0,minimum size=4mm}]
      \draw (1,2) node [regular] {1};
      \draw (2,2) node [regular] {2};
      \draw (1,1) node [regular] {3};
      \draw (2,1) node [regular] {4};
      \draw [<->] (1.15,2) to (1.85,2);
      \draw [->] (1.90,1.90) to (1.10,1.10);
      \draw [->] (1.15,1) to (1.85,1);
      \draw [->] (2,1.15) to (2,1.85);
    \end{tikzpicture}

    \begin{verbatim}
    \end{verbatim}

  \end{center}
  \end{multicols}

  \caption{\label{fig:graph} One positive and one negative example of
  Hamiltonian graphs. On the left is the ILASP representation of the example,
  and on the right is the corresponding graph.}
\end{figure}
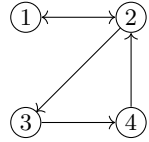

Consider the two context-dependent examples in Figure~\ref{fig:graph}. Both
examples have empty inclusions and exclusions. In the case of a positive
example, this simply means that there must exist at least one answer set of
$B\cup H \cup C$ -- any answer set is consistent with the empty partial
interpretation -- and in the case of a negative example, it means that
there should be no answer set of $B\cup H\cup C$. Given a sufficient number
of examples of this form, ILASP can be used to learn a program that
corresponds to the definition of a Hamiltonian graph; i.e.\ the program
$B\cup H \cup C$ is satisfiable if and only if the set of facts $C$
represents a Hamiltonian graph.  The full program learned by ILASP is:

{\small
\begin{verbatim}
0 { in(V0, V1) } 1 :- edge(V0, V1).
reach(V0) :- in(1, V0).
reach(V1) :- reach(V0), in(V0, V1).
:- not reach(V0), node(V0).
:- V1 != V2, in(V0, V2), in(V0, V1).
\end{verbatim}}

This example shows the high expressive power of ILASP, compared to many other
ILP systems, which are only able to learn definite logic programs. In this
case, ILASP has learned a choice rule, constraints and a recursive definition
of reachability. The full learning task, \texttt{hamilton.las}, used to learn
this program is available online.\footnote{For instructions on how to install
ILASP, see \url{http://www.ilasp.com}. All learning tasks discussed in this
section are available at~\url{http://www.ilasp.com/research}.}

\subsection{Ordering Examples}

Positive and negative examples can be used to learn any ASP program consisting
of normal rules, choice rules and hard constraints.\footnote{This result holds,
up to strong equivalence, which means that given any such ASP program $P$, it
is possible to learn a program that is strongly equivalent to
$P$~\cite{AIJ17}.} As positive and negative examples can only express what
should or should not be an answer set of the learned program, they cannot be
used to learn weak constraints, which do not affect what is or is not an answer
set. Weak constraints create a preference ordering over the answer sets of a
program, so in order to learn them we need to give examples of this preference
ordering -- i.e.\ examples of which answer sets should be preferred to which
other answer sets.  These \emph{ordering examples} come in two forms:
\emph{brave orderings}, which express that at least one pair of answer sets
that satisfy the semantic properties of a pair of positive examples are ordered
in a particular way; and \emph{cautious orderings}, which express that every
such pair of answer sets should be ordered in that way.

Consider a scenario in which a user is planning journeys from one location to
another. All journeys consist of several legs, in which the user may take
various modes of transport. Other known attributes of the journey legs are the
distance of the leg, and the crime rating of the area (which ranges from 0 --
no crime -- to 5 -- extremely high). By offering the user various journey
options, and observing their choices, we can use ILASP to learn the preferences
the user is using to make such choices. The options a user could take can be
represented using context-dependent examples. Four such examples are shown
below. Note that the first argument of the example is a unique identifier for
the example. This identifier is optional, but is needed when expressing
ordering examples.

\begin{verbatim}
#pos(eg_a, {}, {}, {              #pos(eg_c, {}, {}, {
  leg_mode(1, walk).                leg_mode(1, bus).
  leg_crime_rating(1, 2).           leg_crime_rating(1, 2).
  leg_distance(1, 500).             leg_distance(1, 400).
  leg_mode(2, bus).                 leg_mode(2, bus).  
  leg_crime_rating(2, 4).           leg_crime_rating(2, 4).
  leg_distance(2, 3000).            leg_distance(2, 3000).
}).                               }).

#pos(eg_b, {}, {}, {              #pos(eg_d, {}, {}, {
  leg_mode(1, bus).                 leg_mode(1, bus).
  leg_crime_rating(1, 2).           leg_crime_rating(1, 5).
  leg_distance(1, 4000).            leg_distance(1, 2000).
  leg_mode(2, walk).                leg_mode(2, bus).
  leg_crime_rating(2, 5).           leg_crime_rating(2, 1).
  leg_distance(2, 1000).            leg_distance(2, 2000).
}).                               }).
\end{verbatim}

By observing a user's choices, we might see that the user prefers the journey
represented by $\asp{eg\_a}$ to the one represented by $\asp{eg\_b}$. This can
be expressed in ILASP using an ordering example:

\begin{verbatim}
#brave_ordering(eg_a, eg_b, <).
\end{verbatim}

This states that at least one answer set of $B\cup H \cup C_a$ must be
preferred to at least one answer set $B\cup H \cup C_b$ (where $C_a$ and $C_b$
are the contexts of the examples $\asp{eg\_a}$ and $\asp{eg\_b}$, respectively,
and $B$, in this simple case, is empty). The final argument of the brave
ordering is an operator, which says how the answer sets should be ordered. The
operator $\asp{<}$ means ``strictly preferred''. It is also possible to use any
of the other binary comparison operators: $\asp{>}$, $\asp{<=}$, $\asp{>=}$,
$\asp{=}$ or $\asp{!=}$. For instance, the following example states that the
journeys represented by $\asp{eg\_c}$ and $\asp{eg\_d}$ should be equally
preferred.

\begin{verbatim}
#brave_ordering(eg_c, eg_d, =).
\end{verbatim}

By using several such ordering examples, it is possible to learn weak
constraints corresponding to a user's journey preferences. For example, the
learning task \texttt{journey.las} (available online) causes ILASP to learn the
following set of weak constraints:

\begin{verbatim}
:~ leg_mode(L, walk), leg_crime_rating(L, C), C > 3.[1@3, L, C]
:~ leg_mode(L, bus).[1@2, L]
:~ leg_mode(L, walk), leg_distance(L, D).[D@1, L, D]
\end{verbatim}

These weak constraints represent that the user's top priority is to minimise
the number of legs of the journey in which the user must walk through an area
with a high crime rating; their next priority is to minimise the number of
buses the user must take; and finally, their lowest priority is to minimise the
total walking distance of their journey.

Note that in the given scenario there is always a single answer set of $B\cup H
\cup C$ for each of the contexts $C$, meaning that brave and cautious orderings
coincide. When $B\cup H \cup C$ may have multiple answer sets, the distinction
is important, and cautious orderings are much stronger than brave orderings,
expressing that the preference ordering holds universally over all pairs of
answer sets that meet the semantic properties of the positive examples.

\subsection{Noisy Examples}

Everything presented so far in this paper assumes that all examples are
correctly labelled, and therefore that all examples should be covered by the
learned program. In real applications, of course, this is often not the case;
examples may be noisy (i.e. mislabelled), and so finding a program that covers
all examples may not be possible, or even desirable (as this might be
overfitting on the examples). In ILASP, each example can be given a
\emph{penalty}, which is a cost for not covering that example. The search for
an optimal learned program now searches for a program that minimises $|H| +
cost$, where $|H|$ is the length of the program and $cost$ is the sum of the
penalties of all examples that are not covered by the learned program. We have
used this approach to noise to apply ILASP to a wide range of real world
problems, including event detection~\cite{ACS18}, sentence
chunking~\cite{ACS18}, natural language
understanding~\cite{chabierski2017machine} and user preference
learning~\cite{ACS18}.

The function $|H| + cost$ is one example of a \emph{scoring function}. Most
systems, including ILASP, come with a built-in scoring function that cannot be
modified, but in recent work~\cite{lawfastlas}, we have developed a new system
that allows the user to define their own scoring function, allowing a custom
(domain-specific) interpretation of optimality.

\newcommand{\simpleinner}[0]{
  \begin{tikzpicture}
    \node [minimum width=2cm] (3) at (0, 0) {Multi-shot ASP};
    \node [minimum width=2cm] (4) at (0, -0.5) {Solver (Clingo)};
  \end{tikzpicture}
}
\newcommand{\ilasponeinner}[0]{
  \begin{tikzpicture}
    \node [minimum width=2cm] (3) at (0, 0) {Multi-shot ASP};
    \node [minimum width=2cm] (4) at (0, -0.5) {Solver (Clingo)};

    \node [draw,thick,minimum width=3cm] (a) at (0, -2.5) {Compute $VS_n$};
    \node [draw,thick,minimum width=3cm] (b) at (0, -4.5) {Compute $PS_n \backslash VS_n$};
    \draw [->] (0,-1.5) to (a.north) {};
    \draw [->] (a.east) [bend left] to (b.east) {};
    \draw [->] (b.west) [bend left] to (a.west) {};
    \node (4) at (0, -5) {};
    \node [minimum width=2cm] (4) at (-1, -3.5) {n=n+1};
    \node [minimum width=2cm] (4) at (0.5, -1.75) {n=0};
  \end{tikzpicture}
}
\newcommand{\ilasptwoinner}[0]{
  \begin{tikzpicture}
    \node [minimum width=2cm] (3) at (0, 0) {Multi-shot ASP};
    \node [minimum width=2cm] (4) at (0, -0.5) {Solver (Clingo)};

    \node [draw,thick,minimum width=3cm] (a) at (0, -2.5) {Compute optimal $H$};
    \node [draw,thick,minimum width=3cm] (b) at (0, -4.5) {Find violation $vr_i$};
    \draw [->] (0,-1.5) to (a.north) {};
    \draw [->] (a.east) [bend left] to (b.east) {};
    \draw [->] (b.west) [bend left] to (a.west) {};
    \node [minimum width=2cm] (4) at (-0.8, -3.5) {$vr_1,\ldots,vr_i$};
    \node (4) at (0, -5) {};
  \end{tikzpicture}
}

\newcommand{\ilaspmeta}[3]{
  \begin{tikzpicture}
    \node (0) at (0, 0) {ILASP#1};
    \node [draw,thick,minimum width=3cm] (1) at (0, -1) {Pre-processor};
    \node [draw,thick,minimum width=3+#3cm] (2) at (0, -2.4-#3/2) {#2};
    \node [draw,thick,minimum width=3cm] (5) at (0, -3.8-#3) {Post-processor};
    \node [minimum width=3cm] (6) at (0, -4.2-#3) {};
    \draw [->] (1) to (2) {};
    \draw [->] (2) to (5) {};
  \end{tikzpicture}
}
\newcommand{\ilaspmetaextra}[3]{
    \scalebox{0.8}{
      \begin{tikzpicture}
        \node [draw,thick,minimum width=6cm] (a) at (0, 0) {\ilaspmeta{#1}{#2}{#3}};
        \draw [->] (0, 4+#3/2) to (a.north) {};
        \node (e) at (1.2, 3.15+#3/2) {\begin{tikzpicture}
          \node (i) at (0,0) {Task};
          \node (j) at (0,-0.4) {$\langle B, M, E\rangle$};
        \end{tikzpicture}};
        \draw [->] (a.south) to (0, -4-#3/2) {};
        \node (f) at (1.2, -3.15-#3/2) {\begin{tikzpicture}
          \node (g) at (0, 0) {Learned};
          \node (h) at (0, -0.4) {Program ($H$)};
        \end{tikzpicture}};
      \end{tikzpicture}
    }
}

\newcommand{\relfinder}[0]{
  \begin{tikzpicture}
    \node (0) at (0, 0) {Relevant Example Search};
    \node [draw,thick,minimum width=3cm] (1) at (0, -1) {Pre-processor};
    \node [draw,thick,minimum width=3cm] (2) at (0, -2.4) {\begin{tikzpicture}
      \node [minimum width=2cm] (3) at (0, 0) {Single-shot ASP};
      \node [minimum width=2cm] (4) at (0, -0.5) {Solver (Clingo)};
    \end{tikzpicture}};
    \node [draw,thick,minimum width=1.5cm] (6) at (1, -3.8) {Extract $re_i$};
    \node [minimum width=3cm] (7) at (0, -4.2) {};
    \draw [->] (1) to (2) {};
    \draw [->] (2) to (6) {};
  \end{tikzpicture}
}

\newcommand{\hypsearch}[0]{
  \begin{tikzpicture}
    \node (0) at (0, 0) {Program Search};
    \node [draw,thick,minimum width=3cm] (1) at (0, -1.4) {\begin{tikzpicture}
      \node [minimum width=2cm] (2) at (0, 0) {Single-shot ASP};
      \node [minimum width=2cm] (3) at (0, -0.5) {Solver (Clingo)};
    \end{tikzpicture}};
    \node [draw,thick,minimum width=3cm] (4) at (0, -2.8) {Post-processor};
    \node [minimum width=3cm] (5) at (0, -3.2) {};
    \draw [->] (1) to (4) {};
  \end{tikzpicture}
}

\newcommand{\translator}[0]{
  \begin{tikzpicture}
    \node (0) at (0, 0) {Example Translator};
    \node [draw,thick,minimum width=3cm] (1) at (0, -1) {Translate};
    \node [draw,thick,minimum width=3cm] (2) at (0, -2) {Implication Check};
    \node [minimum width=3cm] (3) at (0, -2.4) {};
    \draw [->] (1) to (2) {};
  \end{tikzpicture}
}

\section{Evolution of the ILASP system}

Although we refer to ILASP as a single system, in reality it is a collection of
algorithms, with each algorithm developed to address a scalability weakness of
its predecessor.\footnote{The learning framework has also been expanded with
each new algorithm; however, older algorithms have been updated so that every
ILASP algorithm supports the most general version of the learning framework.}
Table~\ref{tbl:scalability} considers various ``dimensions'' of learning tasks
and shows which ILASP algorithms scale with respect to each of these dimension.

\begin{table}
  \begin{center}
    \scalebox{0.9}{
    \begin{tabular}{|l|c|c|c|c|}
      \hline
                & \multicolumn{4}{c|}{Scales with} \\
      Algorithm & Any Negative Examples? & \# of Examples & Level of Noise & Size of $S_M$
      \\\hline\hline
      ILASP1 & No & No & No & No
      \\\hline
      ILASP2 & Yes & No & No & No
      \\\hline
      ILASP2i & Yes & Yes & No & No
      \\\hline
      ILASP3 & Yes & Yes & Yes & No
      \\\hline
    \end{tabular}
  }
  \end{center}
  \caption{\label{tbl:scalability} A summary of the scalability of each ILASP
  system, with respect to various dimensions of learning tasks.}
\end{table}

Note that although newer versions of ILASP scale with respect to various
dimensions of learning tasks, none of the current ILASP systems scales with
respect to large rule spaces; however, this is being addressed in current work
(for more details, see the next section).

\begin{figure}[t]
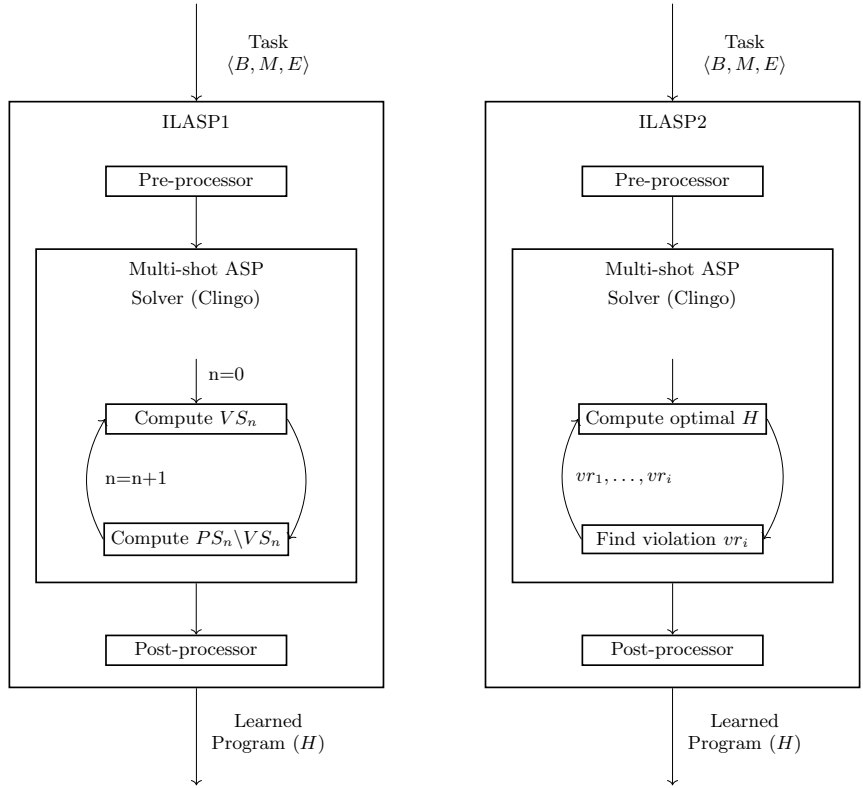

  \begin{multicols}{2}
  \begin{center}
    \ilaspmetaextra{1}{\ilasponeinner}{5}

    \ilaspmetaextra{2}{\ilasptwoinner}{5}
  \end{center}
  \end{multicols}
  \caption{\label{fig:ilasp2} ILASP1 and ILASP2. $PS_n$ and $VS_n$ denote,
  respectively, the positive and violating hypotheses of length $n$ and
  $vr_1,\ldots,vr_i$ are the current violating reasons.}
\end{figure}

\subsection{ILASP1 and ILASP2}

As depicted in Figure~\ref{fig:ilasp2}, the first two ILASP systems both have
three main phases: (1) pre-processing; (2) solving; and (3) post-processing. In
the first phase, ILASP1 and ILASP2 map the input learning task into an ASP
program. Next, this ASP program is solved by the Clingo ASP
solver~\cite{gebser2011potassco,gebser2016theory}. Finally, the answer set
returned by Clingo is post-processed to extract the learned program.

The procedures of the ILASP1~\cite{JELIAILASP} and ILASP2~\cite{ICLP15}
algorithms are encoded using Clingo's built-in scripting feature, which allows a
technique called \emph{multi-shot} solving~\cite{gebser2019multi} to be used.
Multi-shot solving enables a program to be solved iteratively, each time adding
new parts to the program (or removing existing ones). The difference between
the first two ILASP algorithms is in the multi-shot procedure. Both systems
rely on the concepts of \emph{positive hypotheses} -- programs that cover all
of the positive examples -- and \emph{violating hypotheses} -- which also cover
all of the positive examples, but do not cover at least one negative example.
Starting at length $n=0$, in each iteration, ILASP1 computes all the violating
hypotheses of length $n$, converts each violating hypothesis to a constraint,
which is added to the program, and then searches for a positive hypothesis of
length $n$ that does not violate any of the computed constraints (i.e.\ a
program of length $n$ which covers all of the examples). If there is such a
program, it is returned; otherwise, $n$ is incremented and the next iteration
begins. As there can be a large number of violating hypotheses, and because
there is one constraint per violating hypothesis, this process can be very
inefficient if there is at least one negative example.
ILASP2, on the other hand, computes a single (optimal) positive solution $H$ in
each iteration. If $H$ is a violating hypothesis, then it extracts a
``violating reason'' $vr$, which explains why $H$ is violating. It then encodes
$vr$ into a set of ASP rules, which are added to the program in order to rule
out not only $H$, but also any other program which is violating for the same
reason $vr$. Compared with ILASP1, ILASP2 adds far less to the program in each
iteration, and often requires fewer iterations (as the same violating reason
will often apply to many violating hypotheses), leading to orders of magnitude
of improvement in performance on tasks with negative examples.

\subsection{ILASP2i}

The number of rules in the grounding of the ASP encoding used by ILASP1 and
ILASP2 is proportional to the number of examples in the learning task. As the
size of the grounding of an ASP program is one of the major factors in how long
it takes for Clingo to solve that program, this means that ILASP1 and ILASP2 do
not scale with respect to the number of examples.

In real datasets, there is often considerable overlap between the concepts
required to cover several different examples. In other words, there are
\emph{classes} of examples such that each example in a class is covered by
exactly the same programs as every other example in the class. In a non-noisy
setting (where all examples must be covered), only one example per class is
actually required, and all other examples are ``irrelevant''. The idea behind
ILASP2i is to construct a subset of the examples called the \emph{relevant
examples}, which is often significantly smaller than the full set of examples,
but nonetheless still forces ILASP to learn the correct program.
The construction of the relevant examples is achieved by interleaving the
search for an optimal program that covers the (partially constructed) set of
relevant examples with a second search for a new relevant example -- an example
that is not covered by the current program $H$. This interleaving is
illustrated in Figure~\ref{fig:ilasp2i}. At the start of the process, the
program $H$ is set to be empty, because at this point this is the shortest
program that covers the (empty) set of relevant examples. In each iteration,
ILASP2i searches for a new relevant example, and if it finds one, it searches
for a new program (updating $H$) that covers all relevant examples found so
far, using ILASP2 to perform the search. If no relevant example exists, then
$H$ is an optimal solution of the task, and is returned.  An example of the
ILASP2i procedure, based on the coin learning task from the previous section,
is shown in Figure~\ref{fig:ilasp2i-output}.

\begin{figure}[t]
  \begin{center}
    \scalebox{0.6}{
    \begin{tikzpicture}
      \node [draw,thick,minimum width=13cm] (a) at (0, 0) {
        \begin{tikzpicture}
          \node (b) at (0, 4) {ILASP2i};
          \node [draw,thick,minimum width=4cm] (c) at (-4, -1) {\relfinder};
          \node [draw,thick,minimum width=4cm] (d) at (4, -1) {\ilaspmeta{2}{\simpleinner}{0}};
          \node (e) at (0, -5) {};
          \draw [->] (d.north) [bend right] to (c.north) {};
          \draw [->] (c.south) [bend right] to (d.south) {};
          \node (ann1) at (0, -5) {$\langle B, M, \lbrace re_1, \ldots, re_i\rbrace \rangle$};
          \node (ann2) at (0, 3) {Update $H$};
        \end{tikzpicture}
      };
      \draw [->] (-5, 8) to (-5, 1.9) {};
      \draw [->] (-4.5, -1.45) to (-4.5, -8) {};
      \node (i) at (-4.1, 7) {\begin{tikzpicture}
        \node (j) at (0, 0) {Task};
        \node (k) at (0, -0.4) {$\langle B, M, E\rangle$};
      \end{tikzpicture}};
      \node (l) at (-4.3, 3) {$H=\emptyset$};
      \node (f) at (-3.3, -7) {\begin{tikzpicture}
        \node (g) at (0, 0) {Learned};
        \node (h) at (0, -0.4) {Program ($H$)};
      \end{tikzpicture}};
    \end{tikzpicture}
    }
  \end{center}
  \caption{\label{fig:ilasp2i} ILASP2i. The relevant examples
  computed so far are denoted $re_1,\ldots,re_i$.}
\end{figure}
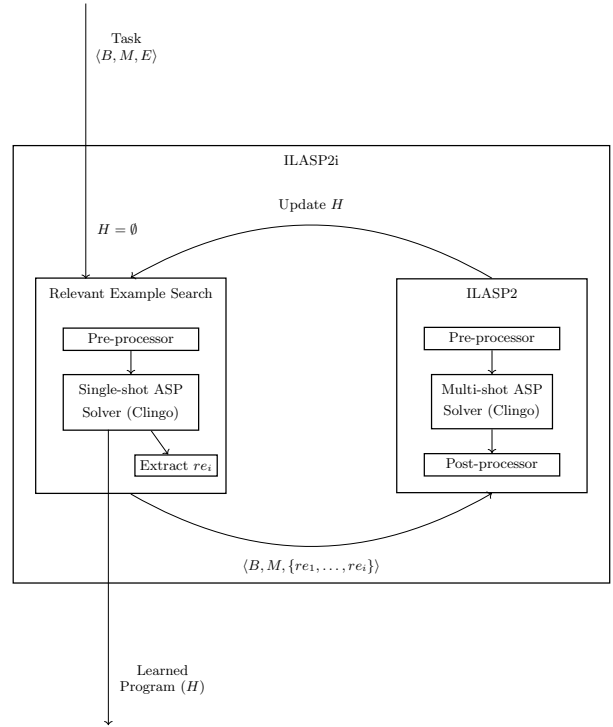

\begin{figure}[t]
  \fontsize{6.5pt}{8pt}
  \begin{multicols}{2}
    \begin{verbatim}
% Background Knowledge

coin(c1).
coin(c2).
coin(c3).

% Examples

#pos(eg1, {heads(c1), tails(c2), heads(c3)},
          {tails(c1), heads(c2), tails(c3)}).

#pos(eg2, {heads(c1), heads(c2), tails(c3)},
          {tails(c1), tails(c2), heads(c3)}).

#pos(eg3, {tails(c1), heads(c2), tails(c3)},
          {heads(c1), tails(c2), heads(c3)}).

#pos(eg4, {tails(c1), tails(c2), tails(c3)},
          {heads(c1), heads(c2), heads(c3)}).


% Mode bias

#modeh(heads(var(coin))).
#modeh(tails(var(coin))).
#modeh(heads(const(coin))).
#modeh(tails(const(coin))).

#modeb(heads(var(coin))).
#modeb(tails(var(coin))).
#modeb(coin(var(coin))).

#constant(coin, c1).
#constant(coin, c2).
#constant(coin, c3).
    \end{verbatim}

    \begin{verbatim}
%%%%%%%%%%%%%%%%%%%%%%%%%%%%%%%%%%%%%%%%%%%%%%%%%
%% Iteration 1
%%
%% Hypothesis:
%%
%% Searching for an uncovered example...
%% The hypothesis does not cover the example: eg1
%%%%%%%%%%%%%%%%%%%%%%%%%%%%%%%%%%%%%%%%%%%%%%%%%
%% Iteration 2
%%
%% Searching for a hypothesis that covers the
%% examples in { eg1 }.
%%
%% Hypothesis:
%%
%% tails(c2).  heads(c1).  heads(c3).
%%
%% Searching for an uncovered example...
%% The hypothesis does not cover the example: eg2
%%%%%%%%%%%%%%%%%%%%%%%%%%%%%%%%%%%%%%%%%%%%%%%%%
%% Iteration 3
%%
%% Searching for a hypothesis that covers the
%% examples in { eg1, eg2 }.
%%
%% Hypothesis:
%%
%% heads(V1) :- coin(V1), not tails(V1).
%% tails(V1) :- coin(V1), not heads(V1).
%%
%% Searching for an uncovered example...
%%%%%%%%%%%%%%%%%%%%%%%%%%%%%%%%%%%%%%%%%%%%%%%%%
%% Solution found:
heads(V1) :- coin(V1), not tails(V1).
tails(V1) :- coin(V1), not heads(V1).
    \end{verbatim}
  \end{multicols}

  \vspace{-5mm}

  \caption{\label{fig:ilasp2i-output} On the left, an extension of the coin
  learning task from Section 2 (with more examples), and on the right, the
  output from ILASP2i. In the first iteration, ILASP2i searches for an example
  that is not covered by the empty program, and finds $\asp{eg1}$. In the
  second iteration, it finds a very specific program that covers $\asp{eg1}$,
  and then finds the second relevant example $\asp{eg2}$. Next, it searches for
  a program that covers both relevant examples, and finds a more general
  program. As this program covers all examples, no further relevant examples
  are computed, and the process terminates. }
  \vspace{-5mm}
\end{figure}

The experiments in~\cite{ICLP16} demonstrate that ILASP2i can be over two orders of
magnitude faster than ILASP2 on tasks with hundreds of examples. The reason is
that the call to ILASP2 calls Clingo with an ASP program whose grounding is
only proportional to the number of relevant examples, rather than to the full
set of examples. Note that although the call to Clingo in the relevant example
search considers all examples, its grounding is not proportional to the number
of examples and it only requires single-shot solving in Clingo, meaning that
each call to Clingo is relatively cheap compared to the Clingo execution in
ILASP2.

\subsubsection{Relation to other approaches.}

ILASP1 and ILASP2 are examples of \emph{batch learners}, which consider all
examples simultaneously. Some older ILP systems, such as
ALEPH~\cite{srinivasan2001aleph}, Progol~\cite{muggleton1995inverse} and
HAIL~\cite{ray2003hybrid}, incrementally consider each positive example in
turn, employing a \emph{cover loop}. The idea behind a cover loop is that the
algorithm starts with an empty program $H$ and, in each iteration, adds new
rules to $H$ such that a single positive example $e$ is covered, and none of
the negative examples are covered. This approach does not work in a
non-monotonic setting, as new rules could ``undo'' the coverage of previously
covered examples. For this reason, most ASP learners are batch learners (e.g.\
\cite{ray2009nonmonotonic,Corapi2012}).
ILASP2i's method of using relevant examples can essentially be thought of as a
non-monotonic version of the cover loop. There are three main differences:

\begin{enumerate}
  \item
    In cover loop approaches, in each iteration a previous program $H$ is
    extended with extra rules, giving a new program $H'$ that contains $H$. In
    ILASP2i, a completely new program is learned in each iteration. This not
    only resolves the issue of non-monotonicity, but is also necessary to
    guarantee that optimal programs are computed. Many cover loop approaches
    make no guarantee about the optimality of the final learned program.
  \item
    In ILASP2i, the set of relevant examples is maintained and used in every
    iteration, whereas in cover loop approaches, only one example is considered
    per iteration.
  \item
    In cover loop approaches, once an example has been processed, even
    if it did not cause any changes to the current program $H$, it is
    guaranteed to be covered by any future program $H'$ and so it is not
    checked again. In ILASP2i, this is not the case. ILASP2i performs the
    search for relevant examples on the full set of examples, even if some were
    previously known to be covered.
\end{enumerate}

ILASP2i is also somewhat similar to \emph{active learning} algorithms, such as
$L^*$~\cite{angluin1987learning}. Active learners are able to query an
\emph{oracle} as to whether what they have learned is correct. The oracle is
then able to provide counterexamples to aid further learning.  ILASP2i's set of
relevant examples are very similar to the counterexamples provided by the
oracle. The main difference between the two approaches is that in active
learning, the oracle is assumed to know the correct definition of the concept
being learned, whereas in ILASP2i, this is not the case, and the search for
relevant examples is only over the provided training examples.

\subsection{ILASP3}

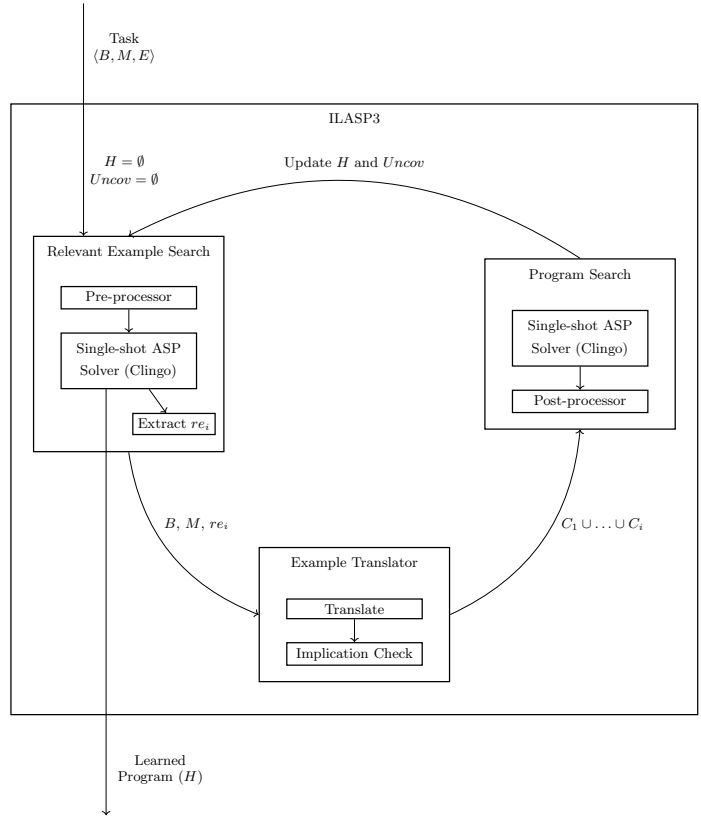
\begin{figure}[t]
  \begin{center}
    \scalebox{0.6}{
    \begin{tikzpicture}
      \node [draw,thick,minimum width=15cm] (a) at (0, 0) {
        \begin{tikzpicture}
          \node (b) at (0, 4) {ILASP3};
          \node [draw,thick,minimum width=4cm] (c) at (-5, -1) {\relfinder};
          \node [draw,thick,minimum width=4cm] (d) at (5, -1) {\hypsearch};
          \node [draw,thick,minimum width=4cm] (t) at (0, -7) {\translator};
          \node (e) at (0, -9) {};
          \draw [->] (c.south) [bend right] to (t.west) {};
          \draw [->] (t.east) [bend right] to (d.south) {};
          \draw [->] (d.north) [bend right] to (c.north) {};
          \node [minimum width=1cm] (ann1) at (-3.5, -5) {$B$, $M$, $re_i$};
          \node [minimum width=1cm] (ann2) at (5.5, -5) {$C_1 \cup \ldots \cup C_i$};
          \node [minimum width=1cm] (ann3) at (0, 3) {Update $H$ and $Uncov$};
        \end{tikzpicture}
      };
      \draw [->] (-6, 9) to (-6, 3.85) {};
      \draw [->] (-5.5, 0.45) to (-5.5, -9) {};
      \node (i) at (-5.1, 8) {\begin{tikzpicture}
        \node (j) at (0, 0) {Task};
        \node (k) at (0, -0.4) {$\langle B, M, E\rangle$};
      \end{tikzpicture}};
      \node (l) at (-5.1, 5.3) {\begin{tikzpicture}
        \node (a1) at (0,0) {$H=\emptyset$};
        \node (a2) at (0,-0.4) {$Uncov=\emptyset$};
      \end{tikzpicture}};
      \node (f) at (-4.3, -8) {\begin{tikzpicture}
        \node (g) at (0, 0) {Learned};
        \node (h) at (0, -0.4) {Program ($H$)};
      \end{tikzpicture}};
    \end{tikzpicture}
    }
  \end{center}
  \caption{\label{fig:ilasp3} The ILASP3 algorithm.}
\end{figure}

Although the concept of relevant examples allows ILASP2i to scale far better
than ILASP2 with respect to the number of examples, this only holds in a
non-noisy setting, where all examples must be covered. When examples can be
noisy, finding a single relevant example only means that a penalty must be paid
if the example is not covered. Many relevant examples (of the same class) may
need to be found before their total penalty makes it ``worth'' covering the
examples. For this reason, the final set of relevant examples is usually much
larger in noisy settings, which significantly reduces the technique's impact on
scalability; in fact, ILASP2i is often even slower than ILASP2 on tasks with
large numbers of potentially noisy examples (i.e.\ large numbers of examples
with finite penalties)~\cite{ilasp_thesis}.

ILASP3 uses a novel method of \emph{translating} an example into a set of
\emph{coverage constraints} over the solution space. The intuition of these
coverage constraints is that they give a list of conditions which are satisfied
by exactly those programs that cover the examples (e.g.\ the learned program
must include at least one of a certain set of rules, or none of another set of
rules). The benefit of having these constraints is twofold. Firstly, finding
the optimal program that conforms to the constraints can be performed using a
single-shot ASP call rather than a multi-shot call, as in ILASP2i, meaning that
in ILASP3 the search for the optimal program is computationally much cheaper.
This is because most of the work is done in the translation procedure (which
does, itself, use a multi-shot call to Clingo). Secondly, once the constraints
for one example have been computed, it is possible to check whether these
constraints are necessary for other examples to be covered. This second benefit
is the main reason why ILASP3 performs so much better than ILASP2i on tasks
with noisy examples. After a relevant example is found (and translated), it is
known that the computed constraints must be satisfied by the learned program,
otherwise a whole set of examples will not be covered (and the penalties of
each example in this set must be paid), rather than just the single relevant
example as in ILASP2i. This can significantly reduce the number of iterations
required by ILASP3, compared with ILASP2i.

Figure~\ref{fig:ilasp3} depicts the procedure of ILASP3. It is similar in
structure to ILASP2i, and similarly interleaves the program search with a
search for relevant examples. The main difference is the addition of the
\emph{Example Translator}, which has two steps: firstly, it translates the
relevant examples into a set of constraints on the solution space; and
secondly, in the \emph{implication check} step, it checks which other examples
would be guaranteed to not be covered if the coverage constraints were not
satisfied -- i.e.\ for which other examples the coverage constraints are
necessary conditions.
%
The coverage constraints give an approximation of the coverage and score of
every program in the program space. This approximation of a program's score is
always guaranteed to be less than or equal to the program's real score, as it
will only overestimate the program's coverage (if the program violates a
coverage constraint, then it is known not to cover the corresponding examples).
The program search computes the optimal program $H$ with respect to the
approximation of the score. When the approximation of the score of $H$ is
correct (i.e.\ the approximation of the coverage of $H$ is equal to the true
coverage of $H$), $H$ is guaranteed to be an optimal solution of the learning
task, and is returned. Checking whether the approximation is correct is
performed by the relevant example search. As the approximation never
underestimates the coverage of $H$, it suffices to only search for a relevant
example within the set of examples that the approximation says $H$ should
cover. To facilitate this, in addition to returning $H$, the program
search also returns the set of examples, $Uncov$, which are known not to be
covered by $H$ (according to the coverage constraints). The relevant example
search is then within $E\backslash Uncov$, rather than the full example set $E$.

In addition to the procedure described in this section, ILASP3 has several
other optional features, designed to boost performance on certain types of
task. For more information, please see~\cite{ilasp_thesis}.

\section{Current and Future Work}

\subsection{Conflict-driven ILP and ILASP4}

Meta-level ILP systems, such as TAL~\cite{Corapi2010}, ASPAL~\cite{Corapi2012}
and Metagol~\cite{muggleton2013meta,metagol}, encode an ILP task as a fixed
meta-level logic program, which is solved by an off-the-shelf Prolog or ASP
solver, after which the meta-level solution is translated back to an
(object-level) inductive solution of the ILP task.

At first glance, the earliest ILASP systems (ILASP1 and ILASP2) may seem to be
meta-level systems, and they do indeed involve encoding a learning task as a
meta-level ASP program; however, they are actually in a more complicated
category. Unlike ``pure'' meta-level systems, the ASP solver is not invoked on
a fixed program, and is instead (through the use of multi-shot solving)
incrementally invoked on a program that is growing throughout the execution.

With each new version, ILASP has shifted further away from pure meta-level
approaches, towards a new category of ILP system, which we call
\emph{conflict-driven}. Conflict-driven ILP systems, inspired by
conflict-driven SAT and ASP solvers, iteratively construct a set of constraints
on the solution space -- where the term constraint is used very loosely to mean
anything that partitions the solution space into one partition that satisfies
the constraint and another that does not -- which must be satisfied by any
inductive solution. In each iteration, the solver finds a program $H$ that
satisfies the current constraints, then searches for a \emph{conflict} $C$,
which corresponds to a reason why $H$ is not an (optimal) inductive solution.
If none exists, then $H$ is returned; otherwise, $C$ is converted to a new
constraint which the next programs must satisfy.

In some sense ILASP2i is already a conflict-driven ILP system, where the
relevant examples in each iteration are the conflicts, although it is not
really in the spirit of a true conflict-driven system as the constraint
generated in each iteration is that one of the examples must be covered, which
was already obvious from the original task. ILASP3 is arguably the first truly
conflict-driven ILP system, as it translates the relevant example (the
conflict) into a set of constraints on the solution space; however, unlike
conflict-driven SAT and ASP approaches the constraints can be extremely large
and expensive to compute, especially when the program space is large. The issue
stems from the fact that the constraints are both sufficient and necessary for
the example to be covered (i.e.\ the example is covered if and only if the
constraints are satisfied). ILASP4, which is currently in development, relaxes
this and computes constraints which are only guaranteed to be necessary (but
may not be sufficient) for the example to be covered. This may mean that the
same relevant example is found twice, leading to more iterations, but each
iteration will be considerably less expensive, and the constraints constructed
in each iteration will be significantly smaller.

\subsection{FastLAS}

Although each ILASP system has improved scalability with respect to several
dimensions, one bottleneck that remains is the size of the rule space.  This is
because every version of ILASP begins by computing the rule space in full.
FastLAS~\cite{lawfastlas} is a new algorithm that solves a restricted version
of ILASP's learning task (currently with no recursion, only observational
predicate learning and no predicate invention). Rather than generating the rule
space in full, FastLAS computes a much smaller subset of the rule space that is
guaranteed to contain at least one optimal solution of the task (called an
\emph{OPT-sufficient subset}). As this OPT-sufficient subset is often many
orders of magnitude smaller than the full rule space, FastLAS is far more
scalable than ILASP. Due to FastLAS's restrictions, once it has computed the
OPT-sufficient subset, it is able to solve the task in one (single-shot) call
to Clingo. FastLAS2, which is currently in development, will lift the
restrictions and replace the call to Clingo with a call to ILASP, thus enabling
ILASP to take advantage of FastLAS's increased scalability.

\bibliographystyle{splncs04}
\bibliography{main}
\end{document}